\documentclass[default,iicol]{sn-jnl}% Default with double column layout
\graphicspath{ {figures/} }

\usepackage{graphicx}
\usepackage{url}
\DeclareUnicodeCharacter{2212}{-}
\jyear{2021}%

\theoremstyle{thmstyleone}%
\theoremstyle{thmstyletwo}%

\theoremstyle{thmstylethree}%

\begin{document}

\title[Article Title]{CNN-BiLSTM model for English Handwriting Recognition: Comprehensive Evaluation on the IAM Dataset}


\author*[1]{\fnm{Firat} \sur{Kizilirmak}}\email{fkizilirmak@sabanciuniv.edu}

\author*[1,2]{\fnm{Berrin} \sur{Yanikoglu}}\email{berrin@sabanciuniv.edu}

\affil*[1]{\orgdiv{Faculty of Engineering and Nat. Sciences}, \orgname{Sabanci University}, \city{Istanbul},  \country{Türkiye}, \postcode{34956}}
\affil*[2]{\orgname{Center of Excellence in Data Analytics (VERIM)}, \city{Istanbul}, \country{Türkiye}, \postcode{34956}}

\abstract{
We present a CNN-BiLSTM system for the problem of offline English handwriting recognition, with extensive evaluations on the public IAM dataset, including the effects of model size, data augmentation and the lexicon. 
Our best model achieves 3.59\% CER and 9.44\% WER using CNN-BiLSTM network with CTC layer.
Test time augmentation with rotation and shear transformations applied to the input image, is proposed to increase recognition of  difficult cases and found to reduce the word error rate by 2.5\% points. 
We also conduct an error analysis of our proposed method on IAM dataset, show hard cases of handwriting images and explore samples with erroneous labels. 
We provide our source code as public-domain, to foster further research to encourage scientific reproducibility.
}


\maketitle

\section{Introduction}\label{sec:intro}

Deep learning models have become the method of choice for handwriting text recognition (HTR) problem, especially in the last decade \citep{Graves-Novel, Shi-CRNN, Bluche-GRCNN, Johannes-EvaluationSeq2Seq, Diaz-Rethinking, Li-TRocr}. 
Most of the recent works in offline handwriting recognition have used Convolutional Neural Networks (CNNs) in combination with recurrent neural networks (RNNs), using Connectionist Temporal Classification (CTC) loss function \citep{Graves-CTC} for training the network. 
This approach allows the model to be end-to-end trainable without requiring explicit image-character alignment.
The CTC algorithm maximizes the total probability of different segmentations over the output of the RNN.
Moreover, the CTC algorithm makes many-to-one alignment in which more than one input time frame could map to a single time frame at the output, allowing characters that do not fit in a single time step to be recognized.

Due to the success of attention based approaches in sequence related areas \citep{Bahdanu-Attention, Vaswani}, more recent works suggested the use of sequence-to-sequence and proposed attention based encoder-decoder models \citep{Johannes-EvaluationSeq2Seq, Kang-Seq2Seq, Kass-AttentionHTR}. 
Among these, earlier methods have combined a CNN with a bi-directional RNN as the encoder and one-directional RNN with attention mechanism as the decoder. More recent approaches utilized Transformer decoders \citep{ Diaz-Rethinking} for decoding or Vision Transformers \citep{Li-TRocr} as a sequence-to-sequence architecture to exploit their applicability over handwriting recognition.

\begin{figure}[t]%
\centering
\includegraphics[width=1.0\columnwidth]{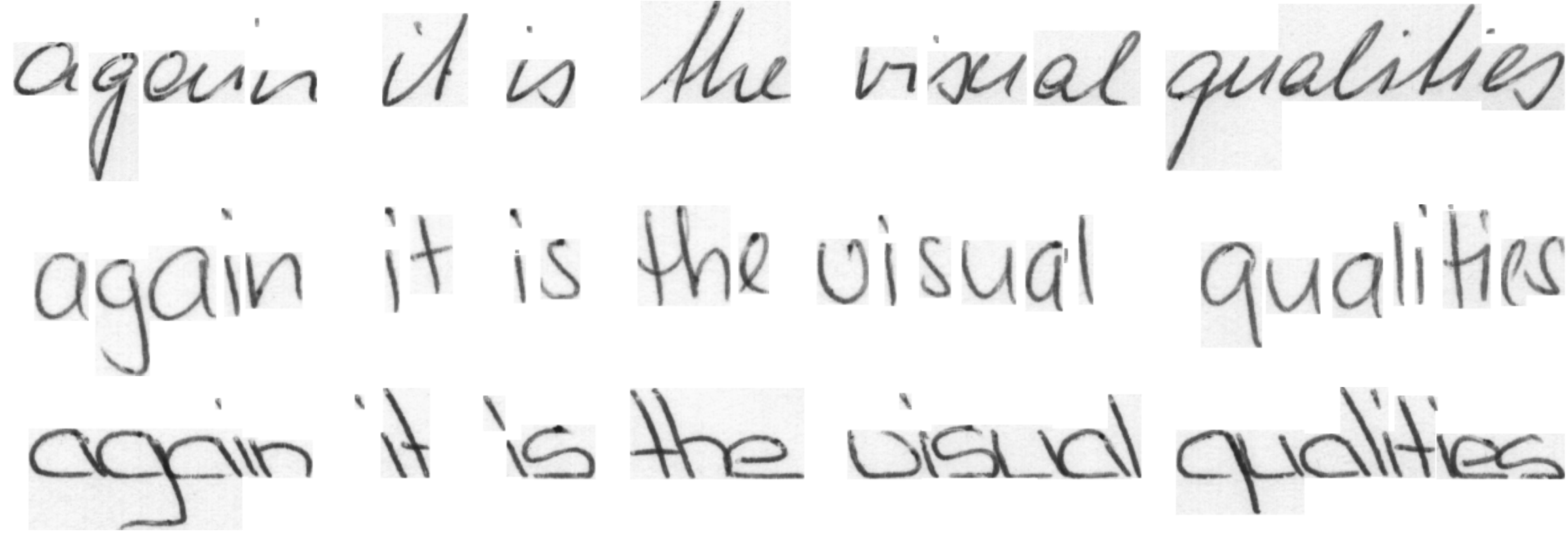}
\caption{The same text written by different people, showing the possible dissimilarities of handwriting. Examples from IAM dataset.}\label{fig:iam-writers}
\end{figure}

Despite recent progress, handwriting recognition technologies fall short in recognizing challenging writing styles. 
In fact, there aren't large public datasets to cover the wide variability in people's handwritings (some examples are given in Figure \ref{fig:iam-writers}). 
This leads us to a data sparseness problem in terms of capturing distinct writing styles. 
In addition, most of the existing datasets comprise historical handwriting images, making it impractical to train and evaluate a model for modern handwriting cases.

There are two common approaches in order to alleviate the data sparseness: augmenting images at train time and generating synthetic handwriting images to pretrain the models \citep{Wigington, Luo-LearnTA, Dutta-Improving, Johannes-EvaluationSeq2Seq, Li-TRocr}. 
In the former, handwriting images are transformed to imitate the same text as if it is written in a different style by changing letter shapes while preserving the readability. 
In the latter case, synthetic handwritten images are created. This approach has the advantage of being able to push the diversity of handwriting styles further.
Both strategies are widely used for introducing different writing styles to increase the generalization capacity of models.

In this paper, we have (1) conducted comprehensive analysis using deep learning models on line-level IAM \citep{IAM} dataset, (2) proposed a simple yet effective test time augmentation method, (3) provided insightful error analysis regarding dataset related issues, (4) explored the state-of-art approaches with their pros/cons and stated directions for future work, (5) made our train, evaluation and benchmarking code public\footnote{https://github.com/firatkizilirmakk/handwriting-recognition} for reproducibility.

The rest of the paper is organized as follows. First the related work regarding the offline handwriting recognition is presented. Then our proposed deep learning model comes, which is followed by data augmentation and synthetic data generation phases, test time augmentation method we proposed, and experiments we performed. Afterwards, error analysis of our model on the IAM dataset and comparison with the state of the art approaches are given.

%%%%%%%%%%%%%%%%%%%%%%%%%%%%%%%%%%%%%%%%%%%%%%%%%%%%%%%%%%%%%%%%%%%%
\section{Related Work}\label{sec:related-work}

\textbf{Hidden Markov Models.} The fundamental approach to handwriting recognition prior to the deep learning era was with Hidden Markov Models (HMMs) \citep{HMM-Survey}. HMMs are doubly statistical models, where there is an underlying stochastic process governing the state transitions and another one for generating observations during state transitions.
Formally the aim is to maximize the posterior probability $P(\mathbf{S} \mid \mathbf{X})$, 
taking into account these stochastic models of state transitions and output generation \citep{Bahl-HMM}.

Preliminary studies explored offline handwriting recognition with a sliding window approach for feature extraction and Hidden Markov Models for generating character or word sequences \citep{Pechwitz-HMM, Bunke-HMM}. They further supported their models with external language models and investigated the effects of lexicon during decoding.\\\\
\textbf{CTC Based Methods.} 
Connectionist Temporal Classification (CTC) method was introduced for speech recognition task \citep{Graves-CTC} in the last decade, allowing RNN models to be trained end-to-end with the backpropagation \citep{Werbos-Backpropagation} algorithm for sequence classification without any need of pre-segmented data. The method was adopted for handwriting recognition \citep{Graves-Novel}.
In CTC based models, (1) a sequence of image features are extracted using a sliding window (1 pixel wide), (2) the extracted features are then fed to a bi-directional LSTM (BiLSTM) \citep{LSTM}, and (3) produced character sequences via the CTC layer. The CTC algorithm takes sequence of probability distributions and generates a character sequence consisting of recognizable characters.  
These models significantly outperformed HMM and HMM-RNN based approaches. Later, Graves et al. applied multi-dimensional LSTM (MDLSTM) layers instead of BiLSTM to incorporate more context around letters and to obtain better transcriptions \citep{Graves-MDLSTM}.

With the rise of deep learning and great performances achieved by CNN models on image processing tasks \citep{Alexnet, Resnet}, researchers have considered deep learning methods for handwriting recognition problem as well. Instead of using hand-crafted image features \citep{Graves-Novel, Graves-MDLSTM}, Shi et al integrated a CNN network to produce more robust image features \citep{Shi-CRNN}. The method first processes input handwriting image with the CNN, generates sequence of image features, and passes them through BiLSTM-CTC layers to obtain final transcription. 
Inspiring from these, Bluche et al proposed a gated convolutional model for computing more generic image features \citep{Bluche-GRCNN}. Puigcerver \citep{Puigcerver-2DLSTM} showed the effectiveness of single dimensional LSTM layers over multi dimensional ones. Further, Dutta et. al. \citep{Dutta-Improving} made a comprehensive study demonstrating effects of data augmentations, pretraining and use of Spatial Transformer Network (STN) \citep{STN}.\\\\
\textbf{Sequence-to-Sequence Approaches.}
The CTC method has the drawback that it prevents the generation of sequences longer than the input sequence, which is a sequence of feature maps \citep{Kang-Seq2Seq, Johannes-EvaluationSeq2Seq}. 
As feature maps get smaller, due to convolution and max pooling operations, the generated sequence becomes shorter, which in turn could result in missing transcriptions. 
Hence, attention based sequence-to-sequence methods have been developed in order to overcome the shortcomings of the CTC and to leverage their sequence learning capabilities on handwriting recognition \citep{Kang-Seq2Seq, Kang-Convolve, Kass-AttentionHTR, Johannes-EvaluationSeq2Seq}. The fundamental idea with these methods is to use CNN-RNN (usually a BiLSTM) to encode the input image as a sequence of features, and then an attention based RNN, usually an LSTM or a GRU, to decode the encoded sequence. The overall network is optimized with the cross entropy loss function, applied over each frame through the output sequence.

Among such approaches, Micheal et al \citep{Johannes-EvaluationSeq2Seq} utilized the CNN-BiLSTM approach where the CNN model learned to encode the handwriting image and the attention-based LSTM model learned to decode the encoded representation into sequence of characters. 
Authors compared different attention mechanisms such as content-based, location-based and penalized attention, and further combined the CTC loss along with the cross entropy loss to increase model capabilities. Kass et al \citep{Kass-AttentionHTR} integrated the spatial transformer network at the forefront of their sequence-to-sequence model to reduce handwriting variations before feeding images to the rest of the architecture. Apart from these, Kang et al \citep{Kang-Seq2Seq} incorporated a character level language model into training phase where they feed the attention based LSTM decoder with the concatenation of encoder and language model outputs.

More recently, transformer \citep{Vaswani} based models have been used, due to their substantial achievements on sequence related tasks \citep{Diaz-Rethinking, Li-TRocr}. Diaz et al compared CTC and sequence-to-sequence approaches, experimented with the transformer decoder and found the best performing model as a self-attention encoder with the CTC decoding. Li et al, on the other hand, employed the Vision Transformer (ViT) \citep{ViT} as an encoder and vanilla transformer decoder, taking advantage of pretrained ViT and transformer decoder. They obtained the state of the art results and further showed the effectiveness of their model and pretraining scheme, without any post processing or external language.\\\\
\textbf{Data Augmentation \& Synthetic Data Generation.} 
In addition to model related developments, most of the studies proposed solutions for dealing with data sparseness issue. There are two common approaches: (1) applying data augmentation techniques at train time for introducing a broad range of handwriting styles, (2) generating synthetic handwriting images to provide enough and diverse handwriting styles to deep learning models.

Affine transformations such as rotating, scaling, and shearing are heavily applied and are shown to be effective methods for mimicking handwriting styles \citep{Poznanski-TTA, Xiao-Rectify, Dutta-Improving}. 
More complex augmentation techniques are also proposed. Wigington et al \citep{Wigington} developed a distortion method along with a profile normalization technique to vary letter shapes, which in turn generates more discriminative letter styles.
Further, Luo et al \citep{Luo-LearnTA} proposed to learn augmentation as a joint-task during training of the networks.

\begin{figure*}[t]%
\centering
\includegraphics[width=0.9\textwidth]{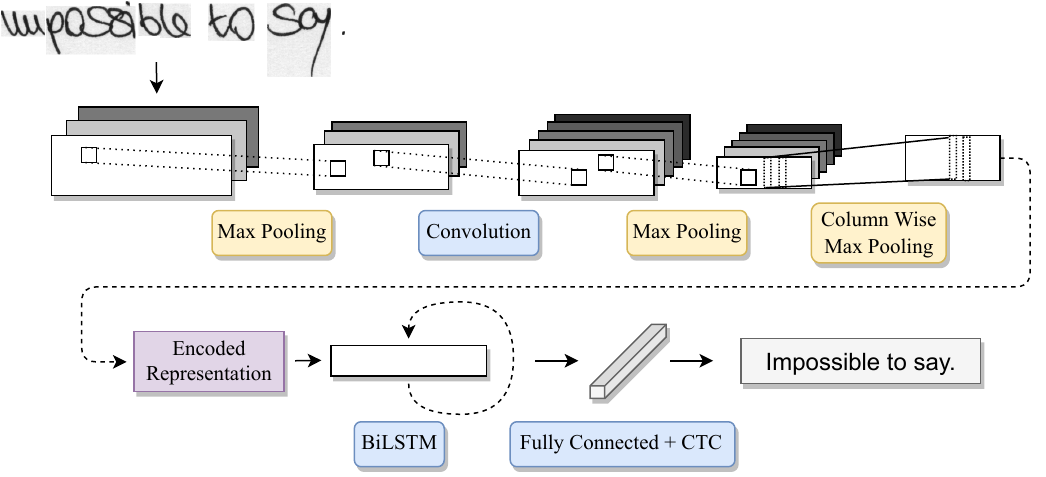}
\caption{Our proposed deep learning network consisting of CNN-BiLSTM models as encoder and CTC as decoder.}\label{fig:model}
\end{figure*}

The second approach to reducing data sparseness involves generating synthetic handwritten samples.
Most of the studies have generated their own synthetic data, either in word or line level, and experimentally shown their effectiveness. While some have synthesized a few millions of handwriting images using words or sentences from known, large corpora \citep{Dutta-Improving, Kang-Seq2Seq, Xiao-Rectify}, others used synthetic data generated for scene-text recognition problem \citep{Kass-AttentionHTR}, or cropped lines from pdf files containing handwriting text \citep{Li-TRocr}. However, though the above methods report performance increase, none of them have published their dataset, which leaves the problem unsolved for other researchers.

As for test time augmentation, there are only a few studies \citep{Poznanski-TTA, Wigington, Dutta-Improving} in which they follow similar approaches. The idea is to augment the input image at the test phase and then generate a new transcription based on the outputs of the original image and of the augmented ones.
Poznanski et al \citep{Poznanski-TTA} applied 36 different transformations on an image at test time, retrieved model outputs for these augmented images and the original one, and afterwards took the mean of these outputs as the final outcome. While Dutta et al \citep{Dutta-Improving} followed the same approach in \citep{Poznanski-TTA}, Wigington et al \citep{Wigington} employed 20 transformations, obtained the corresponding transcriptions and picked the final one with respect to the lowest CTC loss value.
%%%%%%%%%%%%%%%%%%%%%%%%%%%%%%%%%%%%%%%%%%%%% section %%%%%%%%%%%%%%%%%%%%%%%%%%%%%%%%%%%%%%%%%%%%%

\section{CNN-BiLSTM Model}
\label{sec:network}

Inspired from \citep{Shi-CRNN}, and its later accomplished successors \citep{Bluche-GRCNN, Dutta-Improving, Puigcerver-2DLSTM}, we followed a similar deep learning architecture and more details could be found in \citep{FiratThesis}.
The architecture consists of three fundamental blocks: 
feature extraction from the input image using convolutional layers (Section 3.1); 
processing the extracted deep features as a sequence using bidirectional LSTMs (Section 3.2);
and producing the sequence of output  characters with the help of CTC decoding (Section 3.3). 
These modules are described in detail below and the network structure is depicted in Figure \ref{fig:model}.
\subsection{Feature Extraction}
The network uses 12 convolutional layers with $3 \times 3$ kernels for feature extraction. Max pooling operation, with 2 × 2 kernel, is applied two times in total; after the first two and following four convolution layer blocks. The convolution layers in these blocks have 32, 64 and 128 number of filters. ReLU \citep{ReLU} activation function and batch normalization \citep{BatchNorm} are applied after each convolution for faster convergence and easier training. The CNN network produces a feature map tensor in the form of $D\times H\times W$ where $D$ corresponds to number of output filters or depth of feature maps; $H$ and $W$ stand for height and width of feature maps, respectively. Then, we apply a max-pooling function over the height dimension and interchange the dimensions, resulting a tensor  of $W\times D$. 

This output represents $D$ dimensional feature vectors for a sequence of length $W$.
More formally, $W$ could be written as $W = [w_1, w_2, ..., w_t]$, indicating a sequence of $t$ time frames where $w_i \in \mathbb{R}^D$ for $i \in \{1, 2, \dots, t\}$. 

We intentionally avoided using deep CNN networks in order not to narrow down the feature maps too much. Otherwise, the CTC algorithm would only allow producing shorter sequences due to its constraints. Nonetheless, we conducted experiments with a different number of convolution and max-pooling layers and even tried well-known image feature extractors like ResNet18 \citep{Resnet} architecture to decide the baseline model.
We explain the outcomes of these experiments in Section \ref{sec:exp-model}.
\subsection{Sequence Encoding}
This phase processes the sequence of features crafted by the feature extraction step. 
The idea is to incorporate sequence learning capabilities of Recurrent Neural Networks to obtain sequences of features representing the input handwritten text better.

Following \citep{Shi-CRNN, Bluche-GRCNN, Puigcerver-2DLSTM, Dutta-Improving}, we utilised a bi-directional LSTM model as well, leveraging learned contexts from both directions. We used two BiLSTM layers having 256 hidden nodes, without any dropout \citep{Dropout} applied. We fed the BiLSTM layers with the delivered sequence of features, $W\times D$, and the BiLSTM produces a new tensor of size  $W\times(2\times K)$, where $K$ stands for the hidden dimension size of the BiLSTM layer which is equal to 256. 
The produced tensor, sequence of feature vectors, holds an encoded representation of the input handwriting image in a feature space.

Additionally, we explored bi-directional Gated Recurrent Unit (GRU) \citep{GRU} based models and conducted experiments with different number of recurrent layers and hidden dimension sizes. Subsection \ref{sec:exp-model} illustrates the effect of these experiments.
\subsection{CTC Decoding}
After passing the input image through the first two phases, feature extraction and sequence learning, the encoded representation of the input is obtained in the following form $W\times(2\times K)$; formally $W = [w_1, w_2, ..., w_t]$ where $w_i \in \mathbb{R}^H$ for $i \in \{1, 2, \dots, t\}$. 
Here, $H$ stands for $2\times K$ or a real value 512 in our case of $K$ being set to 256.

Then, the encoded sequence representation, $W \times (2 \times K)$, is mapped to a sequence of probability distributions over alphabets of recognizable characters, resulting in $W\times C$. 
To do so, we utilised two fully connected layers; the first one is followed by a ReLU activation and the second one outputs through $C$ number of cells. 
Further, the SoftMax function is applied to produce a sequence of probability distributions defined over the alphabet. Then, CTC function processes the output for labelling the sequence, allowing the whole network to be end-to-end trainable.

\subsection{Word Level Decoding}

We employed three different decoding methods at the test phase to obtain the final transcription from the model output, which is a sequence of probability distributions: greedy, beam search \citep{Graves-beamsearch} and word beam search \citep{WBS}. 
Greedy decoding is the simplest one which picks the most likely character at each time frame. Although it is intuitive to select a character with the highest probability, it is not guaranteed to generate the most likely transcription sequence.

Instead of selecting one character at a time, beam search keeps \textit{k} most likely transcriptions at each time frame and produces a transcription with the highest score among these $k$ alternatives \citep{Graves-beamsearch}. Word Beam Search (WBS), on the other hand, is an extension of the beam search, which constrains words to be in a given lexicon and allows non-word characters to occur inside words \citep{WBS}. Moreover, the algorithm incorporates a 2-gram word-level language model for scoring words through decoding. 
We used WBS algorithm to output the final transcription; set beam width as 150 and configured word beam search to utilize the 2-gram language model.

%%%%%%%%%%%%%%%%%%%%%%%%%%%%%%%%%%%%%%%%%%%%% section %%%%%%%%%%%%%%%%%%%%%%%%%%%%%%%%%%%%%%%%%%%%%
\section{Data Augmentation \& Synthetic Data Generation}\label{sec:aug}

\subsection{Public-Domain Datasets}
The handwriting recognition domain  lacks  high-quality, open-access datasets with a high number of samples, even though the problem has been studied for decades. There are a few  common datasets that contain full sentences; namely, IAM, Rimes, Bentham and Washington \citep{IAM, RIMES, BENTHAM, WASHINGTON} where the last two consist of only historical handwriting images, which are not suitable for modern cases.

IAM dataset was first published in 1999 \citep{IAM}, and still is the most commonly used English handwriting dataset. It contains handwriting texts in English written by 657 different people. There are 1,539 scanned pages, partitioned into 10,373 labelled text lines and 79 different characters. To the best of our knowledge, there is no other public dataset in English with non-historical and line-level handwriting images.

Rimes dataset \citep{RIMES} consist of French handwriting and has been used for evaluating generalization abilities of models for a long time. However, we have not been able to access to the dataset using official channels and their websites, as also indicated here \footnote{https://github.com/jpuigcerver/Laia/issues/44}. Therefore, our experiments will not include results on this dataset.

Due to this data sparseness, current methods usually exploit data augmentation and synthetic data generation techniques. Fundamental idea is to alleviate the lack of data problems and  present various handwriting styles for better generalization. Yet, these works have not published the synthetic datasets they have crafted, which in turn disallows a common evaluation protocol.

\begin{figure}[t]%
\centering
\includegraphics[width=1.0\columnwidth]{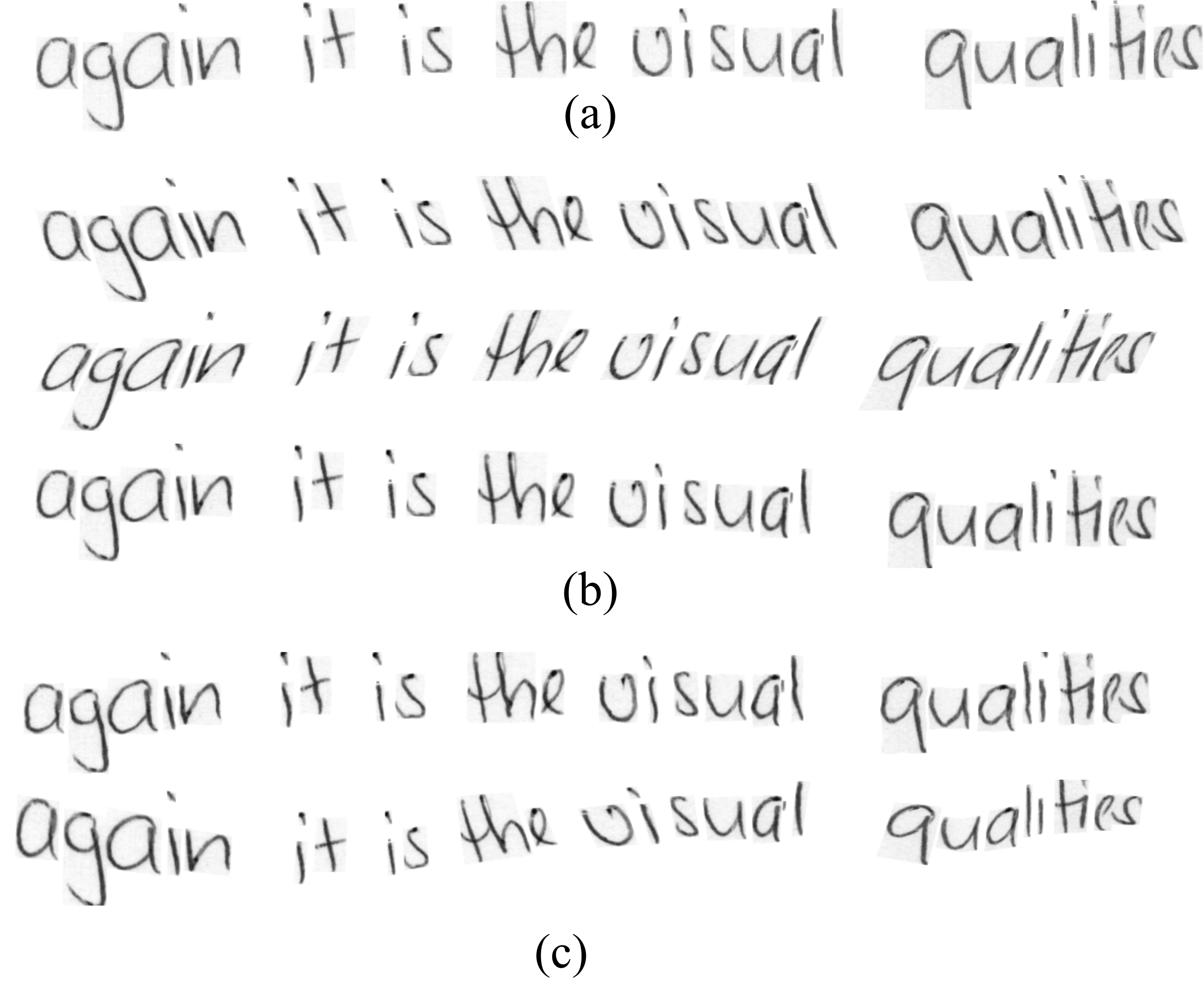}
\caption{Original image along with the augmented samples. (a) Original image. (b) Images augmented by affine transform: first two lines are sheared, the last one is rotated. (c) Images applied elastic distortion and perspective transform from geometric conversions.}\label{fig:augmented-samples}
\end{figure}

\subsection{Data Augmentation}
Although data augmentation for visual tasks has been explored extensively \citep{aug-survey}, there are a couple of studies that proposed novel approaches to augment handwriting images \citep{Simard-Aug, Wigington, Luo-LearnTA}, while others utilized common methods such as affine transformations \citep{Poznanski-TTA, Dutta-Improving, Wigington, Xiao-Rectify}.

Following \citep{Poznanski-TTA, Xiao-Rectify, Dutta-Improving}, we employed affine conversions including shear and rotation to generate natural looking handwriting images.
Moreover, we applied elastic distortion \citep{Wigington} and geometric transformations \citep{Luo-LearnTA} to further increase the handwriting variability.
Augmented examples with these conversions are displayed in Figure \ref{fig:augmented-samples}.

\begin{figure*}[t!]%
\centering
\includegraphics[width=0.7\textwidth]{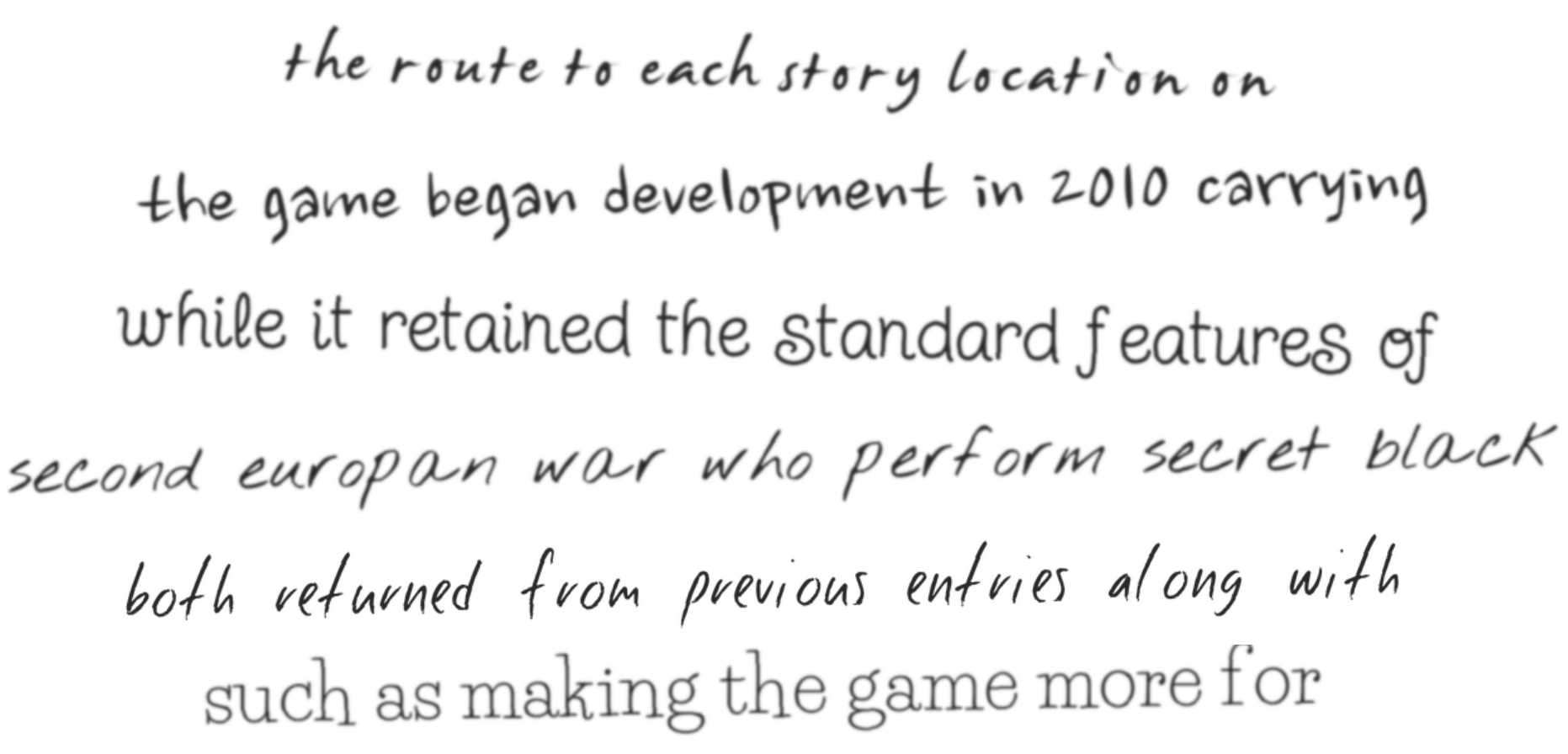}
\caption{Examples of synthetically generated handwriting line images.}\label{fig:synthetic-lines}
\end{figure*}

Shear is a linear affine transformation that takes a point $(x, y) \in \mathbb{R}^2$ and maps it to $(x + ky, y) \in \mathbb{R}^2$, if it conducts horizontal conversion. Here \textit{k} denotes the scaling factor determining how many pixels to move in an image.
Rotation is another linear affine transformation that rotates a point $(x, y)$ in the counter-clockwise direction by $\theta$ degree. Using these shear and rotation transformations makes it possible to mimic slanted handwriting styles as illustrated in Figure \ref{fig:augmented-samples} (b).

Elastic distortion, on the other hand, is a non-linear transformation that elastically changes the shape of a letter. We used the method in \citep{Simard-Aug} where displacement fields $\Delta x$ and $\Delta y$ are sampled from a probability distribution. First, they sample $x$ and $y$ positions of the field from a uniform distribution between $−1, +1$; then, convolve these fields separately with a 1-dimensional Gaussian kernel of $\sigma$ standard deviation; finally, the fields get multiplicated by $\alpha$ value to determine the distortion intensity. Overall, the method is able to distort handwriting letters using $\sigma$ and $\alpha$ parameters.

We also use geometric transformations, as proposed in \citep{Luo-LearnTA}, which transform an image by moving set of predefined points to new positions. The movement of the points is determined via an agent network, where this transformation method is optimized jointly with text recognition module to craft readable yet hard samples. We used closed-form versions of their augmentations\footnote{publicly shared in https://github.com/Canjie-Luo/Text-Image-Augmentation}, namely distortion, stretch and perspective, and grouped them under the name of geometric transformations.

These transformations were applied to the input image more than once with different parameters to generate distinct handwriting styles. 
We decided on the range of parameters of the augmentation methods by visual inspection.
We deformed the letter shapes to some degree while considering the readability of the handwriting. During training, input images undergo a transformation with a probability p = 0.5, which makes it practically impossible to introduce the same image again.

We trained our base model with and without these augmentations in order to evaluate their effects. Outcomes of these experiments are given in detail in Section \ref{sec:exp-aug}.

%
%
%
%%%%%%%%%%%%%%%%%%%%%%%%%%%%%%%%%%%%%%%%%%%%%%%%%%%%%%%%%%%%%%%%%%%%%%%
\subsection{Synthetic Data Generation} 
Many state-of-art systems craft their synthetic dataset following similar steps, such as employing texts written in widely used large corpora \cite{brown-corpus, lob-corpus} or crawling text from the internet \citep{Kang-Seq2Seq} to synthesize artificially written handwritten images.

\begin{figure*}[t]%
\centering
\includegraphics[width=0.7\textwidth]{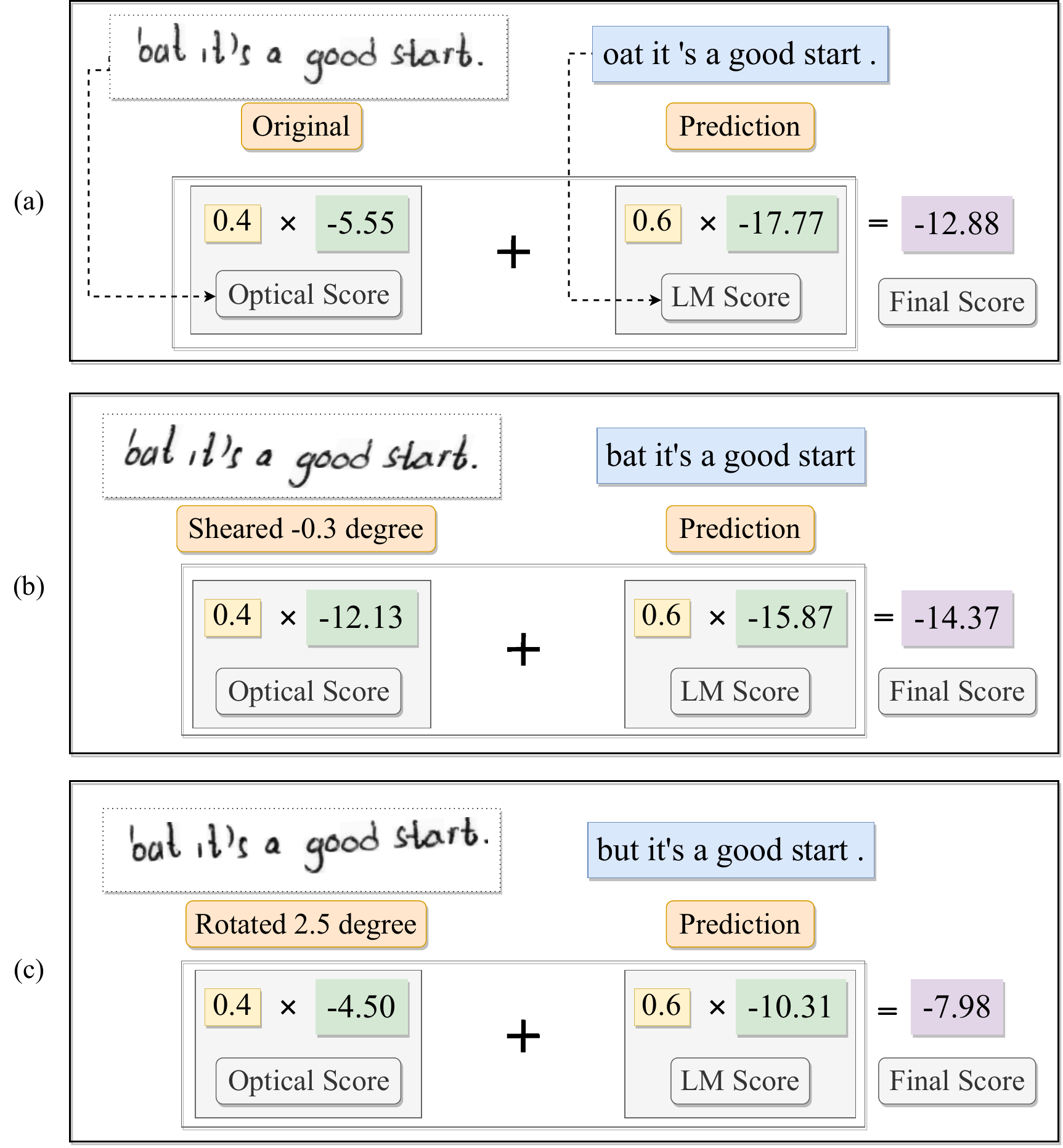}
\caption{An example application of the scoring transcriptions. (a) Output of the original image is scored using the equation \ref{eq:scoring}. The optical score is obtained from the deep learning network. The language model score is computed using the 4-gram language model. (b) and (c) shows the scoring steps of the transformed images. The output with the lowest final score is picked as the final output, which is the correct transcription of the handwriting.}\label{fig:second-decoding-sceheme}
\end{figure*}

We combined WikiText-2 \citep{wikitext-corpus}, Brown \citep{brown-corpus}, and training text of IAM dataset (which corresponds to LOB corpus \cite{lob-corpus}) as the corpus to generate images from. 
Then the texts were preprocessed to have a unified form with correctly tokenized words; sentences were tokenized into words and punctuation was removed.
At most 8 word-sentences are picked to generate lines synthetically.
More than 200 handwriting TrueType fonts were collected from the internet and then utilized to synthesize images. 
We took advantage of the \textit{trdg} tool \footnote{https://github.com/Belval/TextRecognitionDataGenerator} to generate synthetic images using these corpora and fonts. 
Overall, we have generated almost 2.5 million synthetic handwriting image lines, example images are illustrated in Figure \ref{fig:synthetic-lines}. 

%%%%%%%%%%%%%%%%%%%%%%%%%%%%%%%%%%%%%%%%%%%%% section %%%%%%%%%%%%%%%%%%%%%%%%%%%%%%%%%%%%%%%%%%%%%

\section{Test Time Augmentation}\label{sec:tta}
Based on our initial observation that a large majority of character and word errors originated from a small portion of text lines, coupled with the fact that text lines may be recognized better if they were not slanted or rotated,  we hypothesized that test time augmentation would help.

For the offline handwriting recognition problem, the idea of the test time augmentation is to apply transformations to an input handwriting image at the test phase, obtain transcriptions of the corresponding augmented images, and then decide on the best decoding.
The idea was used before in \citep{Poznanski-TTA, Wigington, Dutta-Improving}, however we evaluated different methods for choosing or constructing the final recognition output, given the recognition outputs of different augmentations.
While it is a simple and effective idea, test time augmentation is not widely used or evaluated due to the added decoding time; however, the method could be preferable in the case of batch processing, where time is not too critical.

For test augmentation, we applied $16$ different shear and rotation augmentations in total to the original image at test time and retrieved transcriptions of both augmented images and the original image.
We then obtained the overall score by combining the optical score of the deep learning model, $OP_s$, and a language model score, $LM_s$ over the decoded transcription: 
\begin{equation}\label{eq:scoring}
    score = (\lambda \times OP_s) + (\omega \times LM_s)
\end{equation}
where $OP_s$ is obtained from the WBS decoding algorithm and $LM_s$ denotes the language model score obtained from the below defined 4-gram language model.
Then the transcription with the highest score was picked as the final outcome. Through the experiments, we fit values for $\lambda$ and $\omega$.

We trained the language model as a 4-gram model on WikiText-103 \citep{wikitext-corpus} corpus with Kneser-Ney smoothing \citep{Kneser-Smooth} using KenLM \footnote{https://github.com/kpu/kenlm} tool. Before training the language model, we applied preprocessing steps including tokenization of the words, lowering the letters and removing punctuation.

Figure \ref{fig:second-decoding-sceheme} illustrates the proposed method in a simplified fashion. 
For each image (original or transformed), the optical score, produced by the model for the image and the language model score computed for the corresponding transcription are combined using the above-defined formula. 
The best transcription with the highest score ($-7.98$ for the example in the figure) is picked as the final output, which is the correct decoding obtained from the image rotated by 2.5 degrees for this case.

In addition to the simplicity of this method, it is applicable to any network generating character sequences from handwriting images. 
Further, the method achieves superior results (see Table \ref{table:exp-tta-time}) once the best decoding is selected with respect to the lowest error, as if an oracle telling the truth.
This suggests that a better scoring function will yield greater success.

%%%%%%%%%%%%%%%%%%%%%%%%%%%%%%%%%%%%%%%%%%%%% section %%%%%%%%%%%%%%%%%%%%%%%%%%%%%%%%%%%%%%%%%%%%%

\section{Experiments}\label{sec:experiments}

%In this section, we elaborate on the experiments we have conducted from model development to test time augmentation and more. 
We have run comprehensive evaluations, including the input scaling (\ref{sec:exp-input-scaling}), the baseline architecture (\ref{sec:exp-model}), the data augmentation (\ref{sec:exp-aug}), pretraining with synthetic data (\ref{sec:exp-pretrain}), effect of lexicon, letter case and punctuation (\ref{sec:exp-lexicon}), and the test time augmentation (\ref{sec:exp-tta}).

Through the experiments, models were trained and evaluated on the corresponding Aachen splits of the IAM dataset. Training was set to last by 200 epochs at most and stopped after 10 epochs of non-decreasing validation loss. Initial learning rate was 1e − 3 for most of the experiments though it was tailored in some cases. Rmsprop optimizer was used with weight decay being set to 1e−5. Batch size was kept fixed at 16.

We assessed the performance of the models using character error rate (CER) and word error rate (WER), which corresponds to normalized Levenshtein Distances between the predicted and ground truth character sequences. 
Though we utilized the three decoding methods, we share CER and WER scores from the word beam search algorithm in the tables.

\subsection{Input Scaling} \label{sec:exp-input-scaling}
Input images were resized to $100\times960$ before feeding to recognition models. It is intuitive to keep image height as fixed and resize the width, preserving the aspect ratio. However, in our preliminary experiments we acquired worse performance with this approach and other fixed image sizes. 

We also tried keeping images as is and feeding to models one-by-one as keeping batch size set to 1, yet this approach was neither effective nor efficient.
Thus, we resized images to $100\times960$ and feed through the deep learning network.

%-------------------------------------------------------------------
\subsection{Model Experiments}\label{sec:exp-model}

Our preliminary experiments were designed to decide on the baseline model that would be used in upcoming experiments. Inspired from \citep{Shi-CRNN, Bluche-GRCNN, Puigcerver-2DLSTM, Dutta-Improving}, we followed a similar CNN-RNN-CTC architecture as explained in Section \ref{sec:network}. We explored combinations of CNN and RNN models with different hyperparameters through the experiments.\\\\
\textbf{Backbone.}
We first experimented with the number of convolution layers in the CNN backbone network. Keeping the rest of the layers fixed, we employed separate CNN networks with 8, 10, 12, and 14 layer convolutions while applying 2 or 3 max pooling operations. In addition, a smaller ResNet \citep{Resnet} variant, ResNet18, was utilized to experiment with a deeper model compared to previous ones. 
For the sequence encoding, a two-layered bi-directional LSTM network was used with a hidden dimension of 256. During these backbone and sequence encoding experiments no data augmentation was applied to measure the direct effect of these model choices.

\begin{table}[thb]
\begin{center}
\caption{Impact of backbone configuration on the IAM validation split.}\label{table:backbone}
\begin{tabular}{@{}lccc@{}}
\toprule
Backbone        & \# Max Pooling & CER {\footnotesize \%}    & WER {\footnotesize \%}   \\ \midrule
8 conv. layers  & 2              & 5.49   & 16.34 \\
10 conv. layers & 2              & 5.95   & 17.17 \\
\textbf{12 conv. layers} & \textbf{2}              & \textbf{5.01}   & \textbf{15.14} \\
14 conv. layers & 3              & 5.37   & 15.51 \\ \midrule
%\multicolumn{4}{l}{}                              \\
ResNet18        & default        & 6.08   & 18.27 \\ \bottomrule
\end{tabular}
\end{center}
\end{table}

As displayed in Table \ref{table:backbone}, a 12-layer CNN network with two max pooling operations surpassed the other ones. Though there is no significant performance gap, shallower or deeper the model obtains lower scores; therefore, we preferred continuing with the best scored model with the least parameters and built the baseline model on top of the 12-layer CNN network.\\\\
%
%
%%%%%%%%%%%%%%%%%%%%%%%%%%%%%%%%%%%%%%%%%%%%%%%%%%%%%%%%%%%%%%%%%%%%%%%
%
%
%
%
\textbf{Sequence Encoding.}
After deciding on the backbone network, our next experiment is to select a sequence-encoder model. To this end, we evaluated the performance of bi-directional LSTM and GRU networks. We varied the number of recurrent layers and hidden layer sizes of these deep learning networks. Table \ref{table:seq-encoding} illustrates the model configurations and the corresponding error rates on the validation split of the IAM dataset.

Two recurrent layers fit better for this problem as seen in Table \ref{table:seq-encoding}. Models competed with different hidden dimension sizes yet there was no exact winner of this experiment. However, we intended to keep the model simple while preserving its capabilities. Thus, we preferred to complete the baseline model with 2-layered BiLSTM having hidden dimensions of 256, which outperformed the other networks.

The rest of the experiments were built upon this baseline model consisting of 12 layered CNN with two max pooling functions and two-layered BiLSTM with 256 hidden dimensions followed by a CTC layer.

\begin{table}[thb]
\begin{center}
\caption{Performance of sequence encoding models, BiLSTM and BiGRU, with different configurations on the IAM validation split. Here the backbone network is the one with the 12 convolutional layers and 2 max-pooling operations as displayed in Table \ref{table:backbone}.}\label{table:seq-encoding}
\hspace*{-0.2cm}
\begin{tabular}{@{}ccccc@{}}
\toprule
RNN Model &
  \multicolumn{1}{l}{\# Layers} &
  \multicolumn{1}{l}{\begin{tabular}[c]{@{}l@{}}Hidden\\ Dim. Size\end{tabular}} &
  \multicolumn{1}{l}{CER {\footnotesize \%}} &
  \multicolumn{1}{l}{WER {\footnotesize \%}} \\ \midrule
\multirow{6}{*}{\textbf{BiLSTM}} & \multirow{2}{*}{1} & 256 & 4.95   & 15.42 \\
                        &                    & 512 & 5.26   & 15.99 \\ \cmidrule(l){2-5} 
                        & \multirow{2}{*}{\textbf{2}} & \textbf{256} & \textbf{4.62} & \textbf{14.62} \\
                        &                    & 512 & 4.93   & 14.93 \\ \cmidrule(l){2-5} 
                        & \multirow{2}{*}{3} & 256 & 5.07   & 15.64 \\
                        &                    & 512 & 5.24   & 16.14 \\ \midrule
\multirow{6}{*}{BiGRU}  & \multirow{2}{*}{1} & 256 & 5.27   & 16.54 \\
                        &                    & 512 & 4.90   & 15.48 \\ \cmidrule(l){2-5} 
                        & \multirow{2}{*}{2} & 256 & 5.01   & 15.14 \\
                        &                    & 512 & 4.85   & 14.94 \\ \cmidrule(l){2-5}
                        & \multirow{2}{*}{3} & 256 & 5.05   & 15.53 \\
                        &                    & 512 & 5.70   & 16.33 \\ \bottomrule
\end{tabular}
\end{center}
\end{table}

%%%%%%%%%%%%%%%%%%%%%%%%%%%%%%%%%%%%%%%%%%%%% subsection %%%%%%%%%%%%%%%%%%%%%%%%%%%%%%%%%%%%%%%%%%%%%

\subsection{Data Augmentation}\label{sec:exp-aug}

After establishing the baseline model, we explored the effect of data augmentation techniques, which are described in Section \ref{sec:aug}. 
The transformations were either applied separately or in combination, as shown in Table \ref{table:data-aug}, to assess the effect of each method. 
We only performed one conversion on an image so as not to deform handwritings too much and it was only applied with a probability $p = 0.5$.

\begin{table}[thb]
\begin{center}
\caption{Effect of train time data augmentations applied in separate and in combination using the baseline network defined above. For the combined augmentations, only one of the transformations is used on an image to keep readability of handwritings while deforming handwriting styles. An augmentation is applied with a probability p = 0.5.}\label{table:data-aug}
\begin{tabular}{@{}lcc@{}}
\toprule
Augmentation Method                    & CER {\footnotesize \%} & WER {\footnotesize \%} \\ \midrule
Baseline from Table 2            & 4.62  & 14.62 \\ 
 (No Augmentation)                            & & \\\midrule
Shear                                  & 4.28  & 13.44 \\
Rotate                                 & 4.67  & 14.06 \\
Elastic                                & 4.48  & 14.72 \\
Geometric                              & 4.52  & 14.29 \\
Shear + Elastic + Geometric            & 4.20  & 13.18 \\
Shear + Rotate + Elastic               & 4.17  & 13.21 \\
\textbf{\begin{tabular}[c]{@{}l@{}}Shear + Rotate + Elastic\\+ Geometric\end{tabular}} & \textbf{4.06}  & \textbf{13.02} \\ \bottomrule
\end{tabular}
\end{center}
\end{table}

We determined to parameters of the transformations by visually inspecting augmented images. For the shear transformation, \textit{k} is sampled from a uniform distribution between $−0.6$ and $+0.6$ with a fixed random seed to apply the same conversions through all the experiments. Rotation angle was kept in small degrees from $-2.5$ to $+2.5$ to mimic slanted handwritings while keeping letters inside image borders. Elastic distortion variables were picked from the following sets $\sigma \in \{3, 4\}, \alpha \in \{15,20\}$. Once the parameters set, we trained the models using these augmentations and evaluated without any transformation applied.

Shear conversion is the most convenient way to mimic handwriting thereby the model with only shear augmentation obtained better scores compared to other single transformation  experiments, as revealed in Table \ref{table:data-aug}. 
Even though the other methods were not as effective on their own, their combination achieved better performance due to generating more diverse handwriting styles, as observed over the validation set. 
Hence, we performed the next set of experiments using these augmentations at train time.

%%%%%%%%%%%%%%%%%%%%%%%%%%%%%%%%%%%%%%%%%%%%% subsection %%%%%%%%%%%%%%%%%%%%%%%%%%%%%%%%%%%%%%%%%%%%%
\subsection{Pretraining with Synthetic Data}\label{sec:exp-pretrain}

Generating synthetic handwriting images is another commonly used strategy to reduce the data sparseness and increase the generalization capacity of models. Section \ref{sec:aug} describes in detail how we produced images from the utilized corpora using an online image generation tool.

We, firstly, trained the baseline deep learning network on the generated data consisting of almost 2.5 million synthetic handwriting images. 
We applied the data augmentation methods, explained in Section \ref{sec:aug}, to the synthetic images in order to push the diversity further. 
The synthetic dataset was not split into validation or test partitions; the model was directly trained on the whole dataset by 5 epochs.

\begin{table}[thb]
\caption{Results of our pretrained deep learning model, on the IAM validation and test splits. \textit{Linear layer}: parameters of the output linear layers are updated while fine-tuning. \textit{All layers}: all the parameters are updated.}\label{table:pretrain}
\begin{center}
%\hspace*{-0.65cm}
\begin{tabular}{@{}lcccc@{}}
\toprule
\multirow{2}{*}{Pretraining} & \multicolumn{2}{c}{Validation} & \multicolumn{2}{c}{Test} \\ \midrule
                             & \begin{tabular}[c]{@{}c@{}}CER\\{\footnotesize (\%)}\end{tabular} & \begin{tabular}[c]{@{}c@{}}WER\\{\footnotesize (\%)}\end{tabular}   & \begin{tabular}[c]{@{}c@{}}CER\\{\footnotesize (\%)}\end{tabular}   & \begin{tabular}[c]{@{}c@{}}WER\\{\footnotesize (\%)}\end{tabular}    \\ \midrule
\begin{tabular}[c]{@{}l@{}}Baseline from Table 3\\(No Pretraining)\end{tabular}    & 4.06                     & 13.02                     & 5.20                      & 14.86       \\ \midrule
Pretrained:                                                                                 \\
\hspace{3mm} Linear layer   & 4.21                      & 14.16                     & 5.44                      & 15.58          \\
\hspace{3mm} \textbf{All layers}     & \textbf{3.88}             & \textbf{12.71}            & \textbf{5.05}             & \textbf{14.46} \\ \bottomrule
\end{tabular}
\end{center}
\end{table}

We reduced the character errors by around 0.15\% and word errors around 0.4\% (from 5.20\% to 5.05\% for character errors and from 14.86\% to 14.46\% for word errors) by only pretraining the model on more than 2 million synthetic handwriting images. 
However, there is room for improvement once there is more synthetic and natural handwriting data for pretraining and fine-tuning. 
We performed the next set of experiments using this pretrained model.

Afterwards, the pretrained network was fine-tuned on the IAM dataset train partitions in two ways: (1) freezing all the parameters of the model except the output linear layers. (2) updating all the parameters. Then, the fine-tuned model was evaluated on the IAM validation and test partitions and the corresponding scores are given in Table \ref{table:pretrain}.

Updating only the parameters of the final linear layers did not reveal better scores, 
as expected, due to the difference in handwriting styles between synthetic and natural ones. Hence, we adopted pretraining all layers, in the remainder of the experiments.
The results of this experiment are shown in Table \ref{table:pretrain}.

%%%%%%%%%%%%%%%%%%%%%%%%%%%%%%%%%%%%%%%%%%%%% subsection %%%%%%%%%%%%%%%%%%%%%%%%%%%%%%%%%%%%%%%%%%%%%
\subsection{Effect of Lexicon, Letter Case, and Punctuation}\label{sec:exp-lexicon}
We decoded model outputs using three different methods (greedy, beam search, and word beam search) and gave the best results obtained by the word beam search algorithm in all result tables.
We used greedy and beam search methods as is; in other words, without adding an external language model or considering a lexicon during decoding. 
However, word beam search builds a prefix tree from a corpus i.e. lexicon to keep decoded words in the lexicon. Therefore, we explored the effect of lexicon size and out-of-vocabulary rate on decoding performance.

We experimented with three different English corpora and their combinations for building the lexicon. 
The base lexicon consists of the texts from the IAM training partition, which is part of the LOB corpus \cite{lob-corpus}. 
On top of it, we appended the Brown corpus, which is a million-word English text corpus consisting of texts from 15 different categories. 
We also added the WikiText-2 corpus comprising over two million English words. 
Further, we assessed their combination as well as the validation partition of the IAM dataset.
The lexicon built upon the IAM validation split has zero out-of-vocabulary rate, which is not possible for real-time scenarios yet it underlines the effect of the lexicon.

\begin{table}[thb]
\caption{Reflection of lexicon size with number of unique words (after tokenization, lowering and discarding punctuation) and out-of-vocabulary (OOV.) rates on errors. B stands for the Brown corpus and W2 for the WikiText2. corpus}\label{table: lexicon}
\begin{center}
\begin{tabular}{@{}lcccc@{}}
\toprule
Lexicon    & 
\begin{tabular}[c]{@{}l@{}}\#  \\ Words\end{tabular} &
\begin{tabular}[c]{@{}l@{}}OOV.\\ Rate\end{tabular} & \begin{tabular}[c]{@{}c@{}}CER\\{\footnotesize (\%)}\end{tabular} & \begin{tabular}[c]{@{}c@{}}WER\\{\footnotesize (\%)}\end{tabular} \\ \midrule
\begin{tabular}[c]{@{}l@{}}Baseline from Table 4\\(IAM Train)\end{tabular} & 7.845    & 0.54 & 3.88  & 12.71       \\
\hspace{1.mm}+ B            & 57.606   & 0.15 & 3.49  & 11.77       \\
\hspace{1.mm}+ W2           & 60.244   & 0.21 & 3.56  & 11.92       \\
\hspace{1.mm}\textbf{+ B + W2} & \textbf{82.449}   & \textbf{0.13} & \textbf{3.46}  & \textbf{11.70}       \\
\hspace{1.mm}+ IAM Validation & 2.251  & 0.00 & 2.2   & 6.35        \\ \bottomrule
\end{tabular}
\end{center}
\end{table}

The error rate depends on the context of the words in the lexicon. As the rate of the out of vocabulary (OOV) words decreased, the error rates gets reduced as well, as illustrated in Table \ref{table: lexicon}. In a normal test scenario, it is not possible to contain all likely words in a lexicon, which is the IAM Test case in the table. Nonetheless, most of the time, we obtained lower error rates when using larger lexicons. However, one needs to account the trade-off between the running time with a bigger lexicon and obtained performance. We preferred going with the lexicon consisting of train partition of IAM, Brown and WikiText-2, since there is no significant difference of running time compared to smaller lexicons in Table \ref{table: lexicon}.

The error rate depends on the context of the words in the lexicon. As the rate of the out of vocabulary (OOV) words decreased, the error rates gets reduced as well, as illustrated in Table \ref{table: lexicon}. 
In a normal test scenario, it is not possible to contain all likely words in a lexicon, which is the IAM Test case in the table. 
Nonetheless, most of the time, bigger the lexicon is lower the errors achieved. However, one needs to account the trade-off between the running time with a bigger lexicon and obtained performance. 
We preferred going with the lexicon consisting of train partition of IAM, Brown and WikiText-2, since there was no significant difference of running time compared to smaller lexicons in Table \ref{table: lexicon}.

We decoded and compared the transcriptions with the ground truth text in a case-sensitive manner and also counting punctuation errors. 
However, some work in the literature report that they generated case insensitive text without any punctuation \citep{Wigington, Dutta-Improving}, while others do not specify this. 
As a matter of fact, the IAM dataset contains sentences with several quotes and punctuation around them. As this is an important distinction, we performed experiments considering both strategies, with results given in Table \ref{table:case-punct}.
However, unless otherwise stated explicitly, we shared all the errors measured with case sensitive letters and punctuation through the tables.

\begin{table}[th]
\caption{Effect of the letter case and the punctuation when decoding over the validation split of the IAM dataset. Decoding method up to this experiment involved case sensitive letters with punctuation.}\label{table:case-punct}
\begin{center}
\begin{tabular}{@{}lcc@{}}
\toprule
Case \& Punctuation          & \begin{tabular}[c]{@{}c@{}}CER\\{\footnotesize (\%)}\end{tabular} & \begin{tabular}[c]{@{}c@{}}WER\\{\footnotesize (\%)}\end{tabular} \\ \midrule
Baseline (Case Sensitive + Punctuation)      & 3.46  & 11.70       \\
Case Insensitive + Punctuation               & 3.16  & 10.58       \\
\textbf{Case Insensitive + No Punctuation}   & \textbf{2.87}  & \textbf{8.27}        \\ \bottomrule
\end{tabular}
\end{center}
\end{table}

\begin{table}[th]\caption{Validation and test scores with and without test time augmentation,
in \textbf{case sensitive} recognition, as well as ignoring \textbf{punctuation} mistakes. Oracle indicates the case where the final transcription was selected so as to minimize the  character error, among all transcriptions obtained with test augmentation.}\label{table:tta}
\begin{center}
\begin{tabular}{@{}lcccc@{}}
\toprule
                                                                 & \multicolumn{2}{c}{Validation} & \multicolumn{2}{c}{Test} \\
\begin{tabular}[c]{@{}l@{}}Test Time\\ Augmentation\end{tabular} & \begin{tabular}[c]{@{}c@{}}CER\\{\footnotesize (\%)}\end{tabular} & \begin{tabular}[c]{@{}c@{}}WER\\{\footnotesize (\%)}\end{tabular} & \begin{tabular}[c]{@{}c@{}}CER\\{\footnotesize (\%)}\end{tabular} & \begin{tabular}[c]{@{}c@{}}WER\\{\footnotesize (\%)}\end{tabular} \\ \midrule
-                                                                & 3.46           & 11.70         & 4.80        & 13.85      \\
\checkmark                                                & \textbf{3.22}  & \textbf{10.63} & \textbf{4.37} & \textbf{12.03} \\ \midrule
Oracle                                                           & 2.28           & 8.54          & 3.02        & 9.35       \\ \bottomrule
\end{tabular}
\end{center}
\end{table}

\begin{table}[hb]\caption{Validation and test scores with and without test time augmentation considering \textbf{case insensitive} letters \textbf{ignoring punctuation errors}.}\label{table:tta-uncased}
\begin{center}
\begin{tabular}{@{}lcccc@{}}
\toprule
                                                                 & \multicolumn{2}{c}{Validation} & \multicolumn{2}{c}{Test} \\
\begin{tabular}[c]{@{}l@{}}Test Time\\ Augmentation\end{tabular} & \begin{tabular}[c]{@{}c@{}}CER\\{\footnotesize (\%)}\end{tabular} & \begin{tabular}[c]{@{}c@{}}WER\\{\footnotesize (\%)}\end{tabular} & \begin{tabular}[c]{@{}c@{}}CER\\{\footnotesize (\%)}\end{tabular} & \begin{tabular}[c]{@{}c@{}}WER\\{\footnotesize (\%)}\end{tabular} \\ \midrule
-                                                                & 3.38      & 11.29     & 4.38        & 12.00      \\
\checkmark                                              & \textbf{2.84} & \textbf{9.05} & \textbf{3.59} & \textbf{9.44} \\ \midrule
Oracle                                                  & 1.61          & 6.42          & 2.63        & 7.74       \\ \bottomrule
\end{tabular}
\end{center}
\end{table}

%%%%%%%%%%%%%%%%%%%%%%%%%%%%%%%%%%%%%%%%%%%%% subsection %%%%%%%%%%%%%%%%%%%%%%%%%%%%%%%%%%%%%%%%%%%%%
\subsection{Test Time Augmentation}\label{sec:exp-tta}
Test time augmentation is a simple and effective technique which offers a feasible time complexity for  batch processes. We described our method of using test time augmentation to obtain better transcriptions in Section \ref{sec:tta}.

Once the final transcription is obtained, it is compared with the ground truth character sequence and error score in terms of CER and WER metrics are measured. 
We assessed the proposed test time augmentation method on ignoring errors involving case or punctuation mistakes, to gain further insight. 

Table \ref{table:tta} shows that applying test time augmentation reduced the character errors by almost 0.5\% and word errors by around 2\%, when decoding with case-sensitive letters and punctuation.
The error rate gets decreased further (almost 1\% for character errors and 2.5\% for word errors) when decoding with case insensitive letters without punctuation, as displayed in Table \ref{table:tta-uncased}.

In addition to our scoring function, we performed test time augmentation and selected the final transcription with respect to the lowest CER score, among all transcriptions. We named this method as \textit{Oracle} since it knows in advance the error due to the comparison with the ground truth sequence, which is not a possible in real life. However, our aim was to show the how much gain is possible with this \textit{Oracle} method. As Tables \ref{table:tta},\ref{table:tta-uncased} indicates, there is room for improvement once a better scoring function is employed. Moreover, this strategy is applicable to any model generating character sequences from handwriting images, which makes it considerable for other models as well.

We also analyzed the time 
running times with decoding of original images. 
We run our proposed methods on two different machines with the following configurations: 
(1) AMD Ryzen 3970X, 64 GB ram with Geforce RTX 3090. 
(2) Intel i7 10700K, 32 GB ram with Gefore RTX 3080. 
The time taken for each image was measured and its average with the standard deviation for all the samples in the IAM test partition is computed for both of our methods, as shared in Table \ref{table:exp-tta-time}.
Even though this test time augmentation method slows down obtaining the final transcription, it could be preferred for batch processing of documents when there is no real time requirements. Additionally, better scoring function would reveal greater success without needing any more model parameters or more data.

\begin{table}[th]
\caption{Running times of test time augmentation method on two different machines with CPU \& GPU options. Mean and standard deviation of the elapsed time over the IAM test partition.}\label{table:exp-tta-time}
\begin{center}
\begin{tabular}{@{}llllll@{}}
\toprule
                      &                                                             & \multicolumn{2}{l}{\begin{tabular}[c]{@{}l@{}}Original\\ Time (s)\end{tabular}} & \multicolumn{2}{l}{\begin{tabular}[c]{@{}l@{}}With Test Time\\ Aug. Time (s)\end{tabular}} \\ \midrule
Machine               & \begin{tabular}[c]{@{}l@{}}Processing\\ Device\end{tabular} & Mean                                   & Std.                                   & Mean                                       & Std.                                     \\ \midrule
\multirow{2}{*}{PC 1} & CPU                                                         & 2.58                                   & 1.03                                   & 21.46                                      & 4.08                                     \\
                      & GPU                                                         & 2.27                                   & 0.98                                   & 19.33                                      & 3.84                                     \\
\multirow{2}{*}{PC 2} & CPU                                                         & 3.64                                   & 1.12                                   & 25.54                                      & 4.56                                     \\
                      & GPU                                                         & 3.28                                   & 1.02                                   & 24.76                                      & 4.21                                     \\ \bottomrule
\end{tabular}
\end{center}
\end{table}

%%%%%%%%%%%%%%%%%%%%%%%%%%%%%%%%%%%%%%%%%%%%% subsection %%%%%%%%%%%%%%%%%%%%%%%%%%%%%%%%%%%%%%%%%%%%%

\begin{figure*}[th]%
\centering
\includegraphics[width=0.8\textwidth]{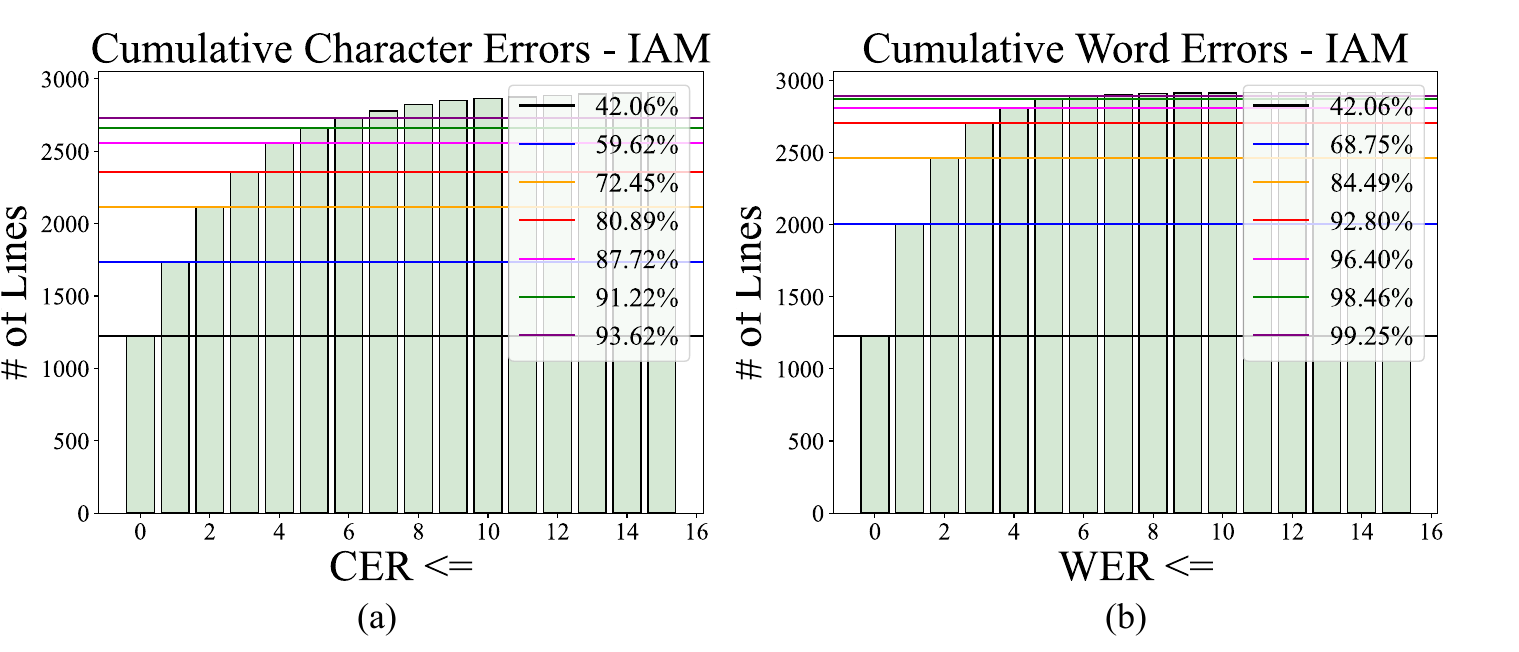}
\caption{Cumulative character and word error rates obtained with the baseline model on the IAM dataset. (a) Illustrates cumulative character errors. Example: All the characters are predicted correctly in 42\% of all the lines, which corresponds to 1226 line images; one or at most two characters are predicted incorrectly in 560 line images, which corresponds to 17\% of all the lines (b) Displays cumulative word error rates following the same manner.}\label{fig:cumulative-error-iam}
\end{figure*}

\section{Error Analysis}
Up to this point, we measured system performance in terms of CER and WER metrics and computed their mean over the IAM test partition. Once we examined the erroneous cases, we found out
that a lot of the errors are contributed by a smaller number of text lines.

In this section,  we provide analysis in terms of the error distribution;  explore difficult samples along with possible ways to overcome these challenging handwriting samples; and share mislabelled example lines from the IAM dataset. We analyzed the outputs of the best models without the test time augmentation, for simplicity.

One can see in Figure \ref{fig:cumulative-error-iam} that  
in 42\% of all the lines, all the characters were recognized correctly and in 80.89\% of all the images in the IAM test set, there were at most 3 character mistakes. Note that the dataset has an average of 35.2 characters per line image with an 8.72 standard deviation. This is a rather surprising statistics, which we believe can indicate certain future research directions. Based on this observation, we focused on test time augmentation, for example.
Similarly, the model produced at most one incorrect word in 68.75\% of all the samples in the test split, where the mean number of words in lines is 8.90. 

\begin{figure*}[t]%
\centering
\includegraphics[width=0.9\textwidth]{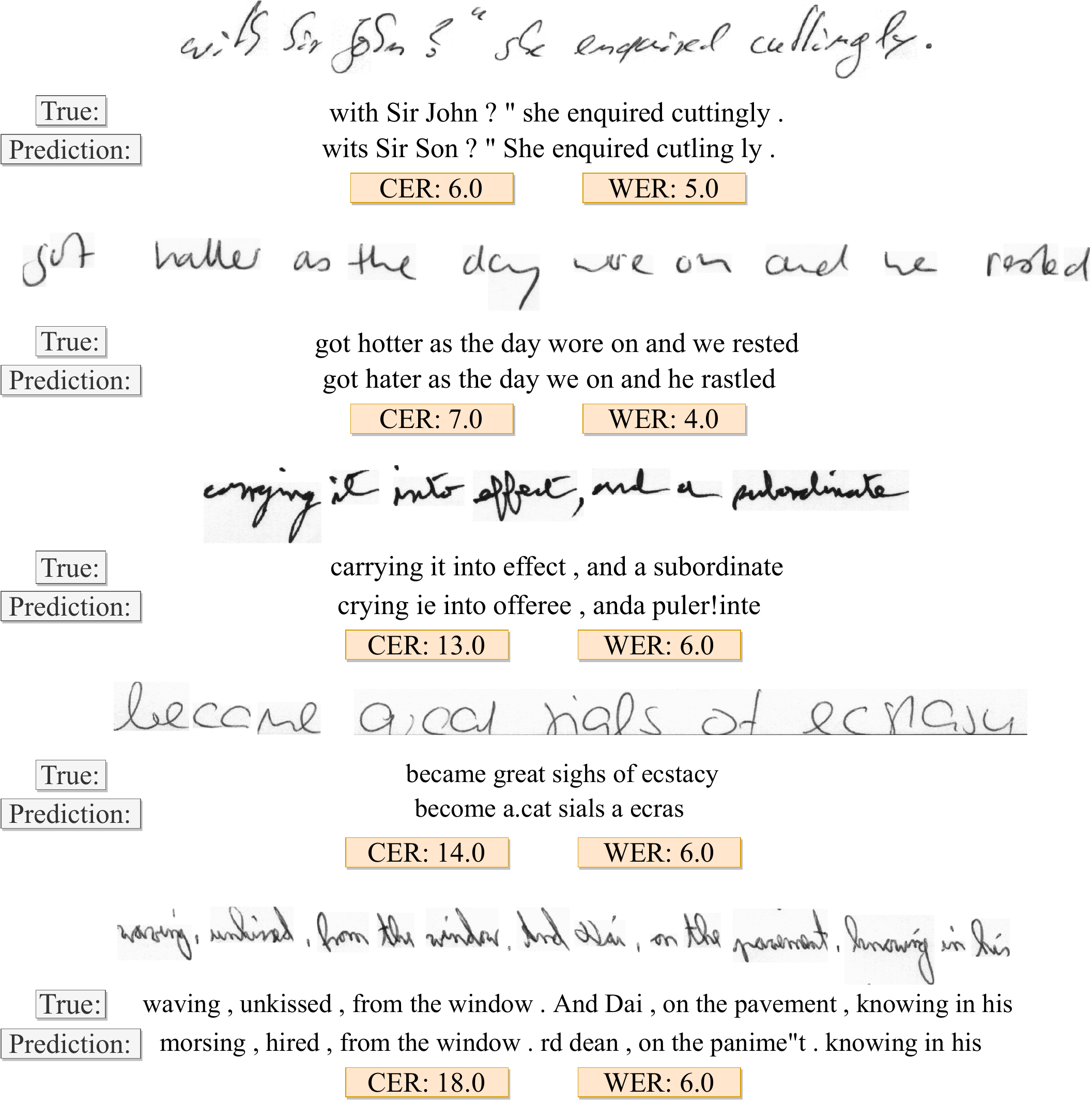}
\caption{Difficult samples from IAM dataset test partition, claiming samples for which model predicted with character errors bigger than 5 i.e. CER $\gt$ 5 are not easy to read by people as well.}\label{fig:iam-most-error}
\end{figure*}

Some of the most challenging writing samples are given in Figure \ref{fig:iam-most-error}. The figure shows examples from IAM test dataset where our model made at least five character errors, which corresponds to 7\% of the test set samples. Note that these handwritings are difficult for people as well.

Furthermore, the IAM dataset contains partially or totally wrong ground truth labels which, in fact, affects the both training and testing phases. Therefore, we found and corrected the erroneous cases on the IAM test split in order to evaluate better. Example erroneous lines along with their issues are presented in the Appendix \ref{sec:appendix}.
We will share the new, fixed metadata of the test split on the link \footnote{https://github.com/firatkizilirmakk/handwriting-recognition}.

%%%%%%%%%%%%%%%%%%%%%%%%%%%%%%%%%%%%%%%%%%%%% section %%%%%%%%%%%%%%%%%%%%%%%%%%%%%%%%%%%%%%%%%%%%%

\begin{sidewaystable*}
\sidewaystablefn%
\centering
\caption{Comparison of our approach with the state-of-the-art approaches. LM \& Lexicon denotes the language model and the lexicon being used (if any) during decoding. 
Decoding indicates which method is used to decode network outputs into character sequence. Here WBS denotes the word beam search algorithm and \textit{bw} is the beam width.
\#P is the number of trainable parameters (in millions) of the deep learning networks. CaseInsens. stands for case-insensitive decoding and NoPunct. denotes the decoding without considering punctuation.}\label{table:comparison}
\footnotesize
\hspace*{-1.3cm}
\begin{tabular}{@{}lcccccccc@{}}
\toprule
Authors & Encoder                  & Decoder                   & Train Set         & LM \& Lexicon   & Decoding   & \#P        & CER \footnotesize\%  & WER \footnotesize\%   \\ \midrule
Bluche \citep{Bluche-GRCNN} & GCRNN                    & CTC                       & IAM + Multi.      & 7-gram LM + 50K & -          & 0.75       & 3.2  & 10.5  \\
Xiao   \citep{Xiao-Rectify} & CNN + LSTM               & CTC                       & IAM + Synth.      & Brown + LOB     & WBS bw/150 & -          & 3.03 & 8.66  \\ \midrule
\vspace{-0.9cm}\\
\multicolumn{9}{l}{}                                                                                                                          \\ \midrule
Michael \citep{Johannes-EvaluationSeq2Seq} & CNN + LSTM               & LSTM w. Att.              & IAM               & -               & Beam bw/16 & -          & 4.87 & -     \\
Kang \citep{Kang-PayAttention}   & \multicolumn{2}{c}{Transformer}                      & IAM + Synth.      & -               & Greedy     & 100        & 4.67 & 15.45 \\
Li \citep{Li-TRocr}     & \multicolumn{2}{c}{Transformer}                      & IAM + Synth.      & -               & Beam bw/10 & 334        & 3.42 & -     \\
Li \citep{Li-TRocr}     & \multicolumn{2}{c}{Transformer}                      & IAM + Synth.      & -               & Beam bw/10 & 558        & 2.89 & -     \\
Diaz \citep{Diaz-Rethinking}    & \multicolumn{2}{c}{\multirow{2}{*}{Self Att. + CTC}} & Public            & 9-gram LM       & Greedy     & $\sim$12.5 & 3.15 & -     \\
Diaz \citep{Diaz-Rethinking}   & \multicolumn{2}{c}{}                                 & Internal + Public & 9-gram LM       & Greedy     & $\sim$12.5 & 2.75 & -     \\ \midrule
\vspace{-0.9cm}\\
\multicolumn{9}{l}{}                                                                                                                          \\ \midrule
\multirow{4}{*}{Ours}   & \multirow{4}{*}{CNN + LSTM}  & \multirow{4}{*}{CTC}  & \multirow{4}{*}{IAM + Synth.} & \multirow{4}{*}{Brown + W102} & WBS bw/150 & \multirow{4}{*}{$\sim$3.6} & 4.80 & 13.85 \\
                        &                              &                       &                               &                               & WBS + TTA  &  & 4.37 & 12.03 \\
                        &                              &                       &                               &                               & \begin{tabular}[c]{@{}c@{}} WBS + TTA +\\CaseInsens. + NoPunct.\end{tabular}  &  & 3.59 & 9.44 \\ \bottomrule
\end{tabular}
\end{sidewaystable*}

%%%%%%%%%%%%%%%%%%%%%%%%%%%%%%%%%%%%%%%%%%%%%%%%%%%%%%%%%%%%%%%%%%%%%%%%%%%%
\section{Comparison with the State of the Art}
Our approach is compared to state-of-art methods with varying characteristics, as shown in  Table \ref{table:comparison}. However, it should be noted that direct comparison of the models may not very meaningful, due to differences or omissions in training or testing criteria.

Our model is able to compete with the similar approaches proposed in \citep{Bluche-GRCNN, Xiao-Rectify}. 
However, we were not able to fully duplicate their results due to the differences including (1) extra data they used in training, (2) the language model they employed to support the deep learning network output, and (3) the lexicon their methods select decoded words from.
Yet, our method achieves a close performance to these works, plus with open and comprehensive evaluation.

Others exploited the attention based approaches to make use of the advantages of these methods on sequence learning. 
In addition to deep learning architectures, they generated millions of synthetic handwriting images and/or crafted real handwriting images through the web, which is another factor of their success. 
As summarized in Table \ref{table:comparison}, Li et. al. \citep{Li-TRocr} pretrained a Transformer model on their synthetic dataset, finetuned over the IAM and achieved superior results without applying any post processing method e.g. a language model or a lexicon. 
Diaz et. al. \citep{Diaz-Rethinking} obtained the state of the art performance where they preferred a simpler model with considerably less amount of parameters, trained on datasets they collected and supported the outputs with a 9-gram language model.

These works suggest that the crucial part of a successful handwriting recognition system is to utilize large number of high quality data, either synthetic or real. Once the data requirement is satisfied, it is better to employ the state of the art deep learning networks such as Transformer, Vision Transformer or other models with attention mechanism. Next, the final output could be decided with the help of an n-gram language model to further increase the success. 

%%%%%%%%%%%%%%%%%%%%%%%%%%%%%%%%%%%%%%%%%%%%% section %%%%%%%%%%%%%%%%%%%%%%%%%%%%%%%%%%%%%%%%%%%%%
\section{Summary and Conclusions}
We proposed a CNN-BiLSTM model for offline English handwriting and comprehensively evaluated it in regards to model architecture, data augmentation, synthetic data generation, and data representation.
We have also suggested a simple yet effective post processing method utilizing test time augmentation. Even though the method takes more time compared to decoding of a single image, it could be preferable in batch processes that do not have time constrains and it is applicable to any model producing sequence of characters from images. 

In addition to the usual metrics for assessing handwriting recognition models, we analyzed the error distributions to gain insights regarding challenges. We have shown that the majority of the errors stem from a relatively small portion of the test cases, which can be instructive for improvements. 

We will share our code openly upon publication, to allow for other researchers, as well as reproducibility.

\bibliographystyle{splncs04}
 \bibliography{ICDAR23}

\newpage
\begin{appendices}
\section{Incorrect IAM Test Samples}\label{sec:appendix}

\begin{figure}[h]%
\centering
\includegraphics[width=0.8\textwidth]{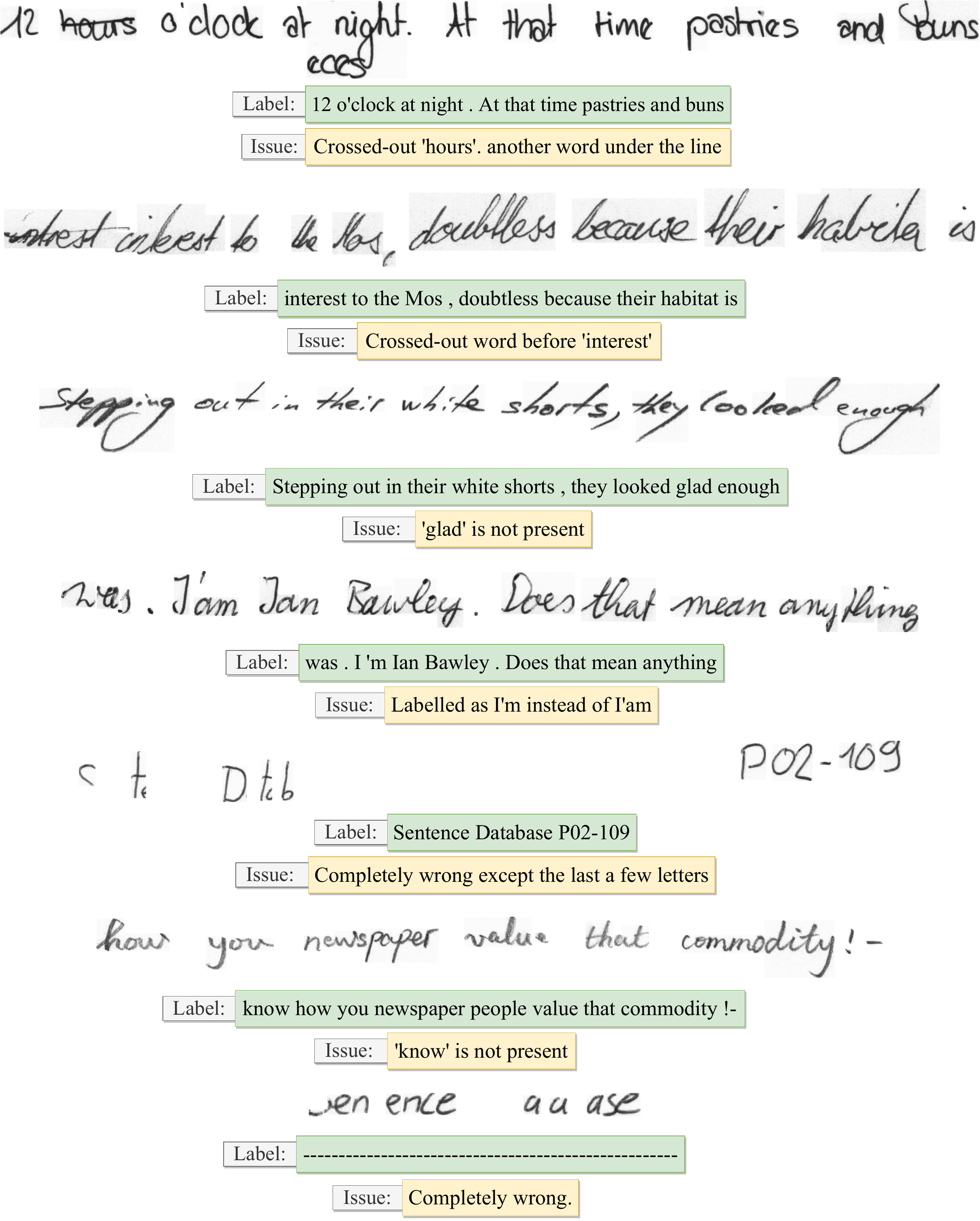}
\caption{Samples from lines in the IAM dataset test partition with wrong labels. \textit{Label} denotes the ground-truth text written on the image and \textit{Issue} indicates the error of the label e.g. a word in a label is not present in the corresponding image. The corrected labels of the IAM test split will be shared upon acceptance.}
\label{fig:iam-incorrect-samples}
\end{figure}

\end{appendices}

 \end{document}